\pdfoutput=1
\documentclass[11pt]{article}

\usepackage[margin=1in]{geometry}
\usepackage{cite}
\usepackage{amsmath,amssymb,amsfonts}
\usepackage{graphicx}
\usepackage{textcomp}
\usepackage{xcolor}
\usepackage{booktabs}
\usepackage{tabularx}
\usepackage{array}
\usepackage{url}
\usepackage{xurl}
\usepackage{authblk}
\usepackage[font=small,labelfont=bf]{caption}
\usepackage[section]{placeins}
\usepackage{wrapfig}
\usepackage[hidelinks]{hyperref}

\hypersetup{
  pdftitle={IO Factory: Simulating AI-Enabled Influence Campaigns at Scale},
  pdfauthor={Lukasz Olejnik, Wenchao Dong, Jonas R. Kunst, Signe Riemer-Sorensen, Tobias Herb, Meeyoung Cha, and Daniel Thilo Schroeder},
  pdfkeywords={AI-enabled influence, AI swarms, computational social systems, information operations, multi-agent simulation, social simulation, LLM agents, exposure measurement, red teaming}
}

\newenvironment{keywords}{%
  \par\noindent\textbf{Keywords: }%
}{\par\vspace{0.5em}}

\begin{document}

\title{\huge\bfseries IO Factory: Simulating AI-Enabled Influence Campaigns at Scale}

\author[1,2]{Lukasz~Olejnik}
\author[3]{Wenchao~Dong}
\author[4,5]{Jonas~R.~Kunst}
\author[6]{Signe~Riemer-S{\o}rensen}
\author[6]{Tobias~Herb}
\author[3,7]{Meeyoung~Cha}
\author[6]{Daniel~Thilo~Schroeder}

\affil[1]{Department of War Studies, King's College London, London, United Kingdom}
\affil[2]{Independent researcher}
\affil[3]{Max Planck Institute for Security and Privacy, Bochum, Germany}
\affil[4]{Department of Communication and Culture, BI Norwegian Business School, Oslo, Norway}
\affil[5]{Department of Psychology, University of Oslo, Oslo, Norway}
\affil[6]{SINTEF Digital, Oslo, Norway}
\affil[7]{School of Computing, Korea Advanced Institute of Science and Technology (KAIST), Daejeon, Republic of Korea}

\date{}

\maketitle
\begingroup
\renewcommand\thefootnote{}
\footnotetext{Corresponding authors: Lukasz Olejnik (\texttt{me@lukaszolejnik.com}) and Daniel Thilo Schroeder (\texttt{daniel.t.schroeder@sintef.no}).}
\endgroup
\begin{abstract}
We introduce IO Factory, an AI-driven framework for simulating information and influence campaigns as fully integrated, traceable processes. The threat of digital manipulation now extends beyond persuasive text from individual language models to AI swarms, i.e., persistent groups of coordinated agents that adapt to platform feedback and disguise organized campaigns as ordinary social interaction. Because such campaigns cannot be identified from isolated messages alone, they must be analyzed across a continuous spectrum of planning, platform action, exposure, interpretation, measurement, and adaptation. IO Factory represents this process inside a controlled simulated platform, linking actor roles, platform actions, exposure records, structured model-based evaluations, and configured changes in the simulated population. We implement the architecture and evaluate it across configurations of up to 100,000 agents. The results show that IO Factory executes campaign timelines at scale and produces inspectable evidence of exposure and measured movement in configured belief variables. By recording the actors, objectives, action constraints, exposure paths, and measurement rules used in each run, IO Factory supports reproducible research and red-team analysis of coordinated influence.
\end{abstract}


\section{Introduction}\label{sec:introduction}
Digital influence campaigns did not begin with generative AI. Earlier operations relied on social bots, troll farms, and coordinated human operators~\cite{ferrara2016rise,bradshaw2019global}. These systems could amplify messages, repeat slogans, and create artificial engagement, but they had clear limits. Bots were often repetitive, brittle, and poor at conversation. Human operators could produce more plausible content, but they were expensive, slow, and difficult to scale. They could coordinate, but they could not easily appear as many different ordinary people over long periods of time.

Generative AI changes this dynamic. Large Language Models (LLMs) make it inexpensive to produce fluent, tailored, and context-aware messages at scale~\cite{schroeder2026swarms,olejnik2025propagandaFactories,orlando2026emergent}. In many ordinary settings, such text may no longer be reliably distinguishable from human-written text by style alone. LLMs can draft replies, translate arguments into local idioms, adjust tone for different communities, and sustain long interactions. At campaign level, these capabilities shift influence operations from periodic bursts of propaganda toward persistent systems that can act over time.

A further shift comes from agentic AI. LLMs can now be connected to memory, tools, and platform accounts. This makes it possible to build AI swarms: groups of agents that keep stable identities and act toward a shared campaign goal~\cite{schroeder2026swarms,orlando2026emergent}. The risk is campaign behavior that appears dispersed at the account level while remaining coordinated at the system level. Different accounts can write in different voices and enter the conversation at different moments while still moving the same narrative or creating the appearance of social consensus.

This creates an observability problem for research and defense. The visible pieces of a campaign may look ordinary: a plausible post, a familiar account, or a real person repeating a message. In cyborg propaganda, human accounts can become the delivery layer for centrally shaped AI content~\cite{kunst2026cyborg}. In AI swarms, artificial accounts can sustain ordinary-looking identities over time. The boundary between organic and engineered behavior, therefore, becomes hard to draw from content or single-account behavior alone; the relevant object is collective campaign behavior across accounts, time, and exposure paths~\cite{pacheco2021uncovering,mannocci2024detection}.

At the same time, the data needed to study these campaigns is often difficult to access. Researchers, journalists, and civil-society organizations may not see the cross-platform, temporal, and relational structure of coordinated campaigns. Platform access is restricted, data is fragmented, and important signals may become clear only after the campaign has unfolded. This means that content detection is not enough. The relevant signal is often behavioral and temporal: coordination, role division, repeated exposure paths, narrative movement through sources, adaptation after feedback, or the gradual construction of synthetic consensus~\cite{graphika2020secondaryinfektion,brookings_abcd}.

This is why simulation and red teaming become necessary~\cite{perez2022red}. The value of simulation is to let researchers represent campaign designs, vary assumptions, and inspect evidence paths under controlled conditions. IO Factory contributes such a setting by making campaign structure, platform contact, exposure, measurement, and comparison inspectable in recorded runs.

Synthesizing existing operational models of information operations, IO Factory treats the campaign lifecycle as the core unit for controlling, measuring, and inspecting information-operation dynamics~\cite{cset2021richdata,carnegie2023phase,disarmframework,microsoft2022cyberinfluence}. In the reported implementation, this lifecycle is instantiated as ten phases, from reconnaissance and narrative design through amplification, adaptation, and evaluation (see Table~\ref{tab:lifecycle-phase-semantics}).

Section~\ref{sec:env_and_actors} defines the three actor categories used in the reported experiments. Civilians form the measured audience. IO operators act through public platform accounts in the active condition. The manager is a non-public control component that can assess the simulation run and issue bounded guidance. Within this controlled setting, IO operators can publish or relay content on a simulated public (social media) platform; platform feeds and discovery mechanisms determine which content becomes visible; and civilians may encounter, interpret, react to, or further circulate that content.

IO Factory measures campaign influence as directional lift: the additional movement of the simulated civilian population toward the configured campaign target in the active run, compared with a matched baseline run in which IO operators are absent. This movement is computed over constructs, meaning simulator-scale audience variables such as trust in a source or support for a policy. A post can contribute to this measure only after it becomes visible to a civilian, enters the exposure record, is evaluated against the relevant construct definition, and passes the update rule. The resulting value is a simulator-scale outcome. It makes exposure, measurement, and comparison explicit for defensive research and red-team evaluation, while leaving external claims to later validation against real-world evidence.

This paper asks whether an AI-enabled influence campaign can be represented as a traceable lifecycle inside a controlled simulation. Campaign guidance is translated into actor actions. Actor actions change platform state. Platform rules create visibility. Visibility produces civilian exposures. Exposures enter measurement, and measured changes are compared against a matched baseline. IO Factory preserves this chain as the unit of analysis. The contribution is therefore about representation, traceability, and controlled comparison at scale.

The paper contributes a lifecycle representation for AI-enabled influence campaigns, an architecture that separates the simulated platform environment from the campaign model under study, a measurement pipeline that follows content from creation through visibility and exposure to deterministic state update, and a matched comparison protocol that keeps campaign activity separate from audience response. It also reports empirical feasibility at scale, including configurations with up to 100,000 agents. Together, these contributions establish IO Factory as a research and red-teaming environment for testing explicit campaign assumptions alongside real-world evidence~\cite{perez2022red,yamin2020cyber}.

The remainder of the paper proceeds as follows. Section~\ref{sec:related_work} reviews related work. Section~\ref{sec:framework_overview} presents the framework architecture. Sections~\ref{sec:env_and_actors}--\ref{sec:measure} describe the simulated environment and actors, campaign lifecycle, and measurement protocol. Section~\ref{sec:simulation-results} reports simulation runs and results. Section~\ref{sec:discussion} discusses limitations and future work. Section~\ref{sec:conclusion} concludes.

\section{Related Work}\label{sec:related_work}
\subsection{LLM-Powered Social Simulation and Multi-Agent Systems}
Social simulation uses language models to populate simulated environments with agents, content, and interaction traces. Prior work shows that LLM agents can maintain memory, plan, reflect, interact through environments, and generate social behavior at scale~\cite{park2023generative,tang2025gensim}. Other studies use language models as simulated participants or agents in opinion dynamics, showing both the promise and the limits of synthetic populations~\cite{aher2023using,chuang2024simulating}. IO Factory builds on this environment-based view of LLM agents, but focuses on intervention studies: lifecycle phase, feed exposure, construct update, provenance, and active-baseline comparison become part of the study record.
\subsection{Information Operations and AI-Enabled Influence}
Research on disinformation, computational propaganda, and coordinated platform manipulation treats influence as organized campaign activity rather than isolated content~\cite{bradshaw2019global,starbird2019disinformation,mannocci2024detection}. The same message can matter differently depending on source, path, repetition, and audience trust. Recent work on AI swarms, propaganda factories, hybrid human-AI activity, and LLM-agent simulations shows how language models can support persistent personas, low-cost generation, shared goals, model-based judging, and coordinated or human-carried narratives~\cite{schroeder2026swarms,olejnik2025propagandaFactories,kunst2026cyborg,orlando2026emergent}. IO Factory complements this work by turning the campaign process into something a researcher can design, vary, measure, and compare under controlled conditions.
\subsection{Campaign Lifecycles, Platforms, and Measurement}
Operational models describe information operations as staged processes, including audience analysis, infrastructure preparation, narrative development, seeding, amplification, engagement, assessment, and adaptation~\cite{cset2021richdata,carnegie2023phase,disarmframework,microsoft2022cyberinfluence}. IO Factory adopts this lifecycle view by making phase state part of the simulation record: a phase controls which actors can act, what can become visible, and what evidence is recorded before the next phase.

Platform and diffusion research motivates separating message production, reach, interpretation, and effect~\cite{bakshy2015exposure,vosoughi2018spread,eady2023exposure}. Misinformation and persuasion research further shows that message volume alone does not determine effect; source trust, repetition, prior belief, familiarity, correction, and social identity can all affect interpretation~\cite{roozenbeek2024beyond,ecker2022psychological,pennycook2021psychology,dechene2010truth,petty1986elaboration,chaiken1980heuristic}. IO Factory turns these factors into explicit simulator parameters, so the conditions under which a message can move simulated state are declared rather than left implicit.

Finally, work on reproducibility, auditing, red-team evaluation, and LLM-as-judge methods explains why the evidence layer matters~\cite{national2019reproducibility,raji2020closing,perez2022red,yamin2020cyber,zheng2023judging}. A simulation result is interpretable only when the comparison is explicit: what changed and what stayed fixed. Since model-based judges are measurement instruments, not ground truth~\cite{zheng2023judging}, IO Factory links campaign lifecycle, platform exposure, judge records, state updates, and matched comparison records in one audit trail.

\section{Framework Overview}\label{sec:framework_overview}

\begin{figure}[!htbp]
  \centering
  \includegraphics[width=\textwidth]{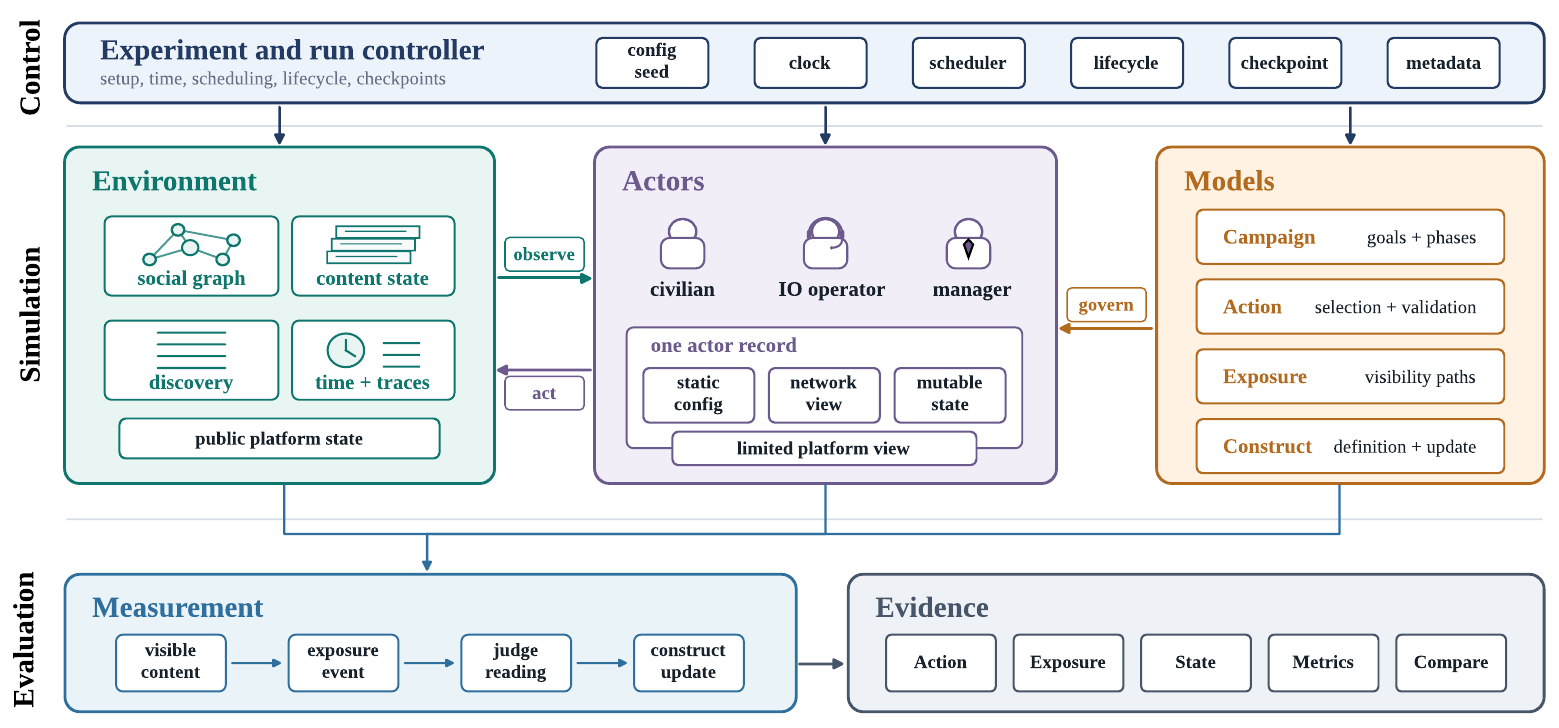}
  \caption{High-level IO Factory architecture. The run controller orchestrates experiment execution across the platform environment, actors, study models, measurement, and evidence layers. Provenance links preserve the path from public action through exposure and state updates to comparison and audit records.}
  \label{fig:system-overview}
\end{figure}

IO Factory is organized around the components shown in Figure~\ref{fig:system-overview}, which arranges six components into three planes: a control plane that steers experiment execution, a simulation plane in which platform activity unfolds, and an evaluation plane that turns this activity into measurements and evidence. This section provides the system map before Sections~\ref{sec:env_and_actors}, \ref{sec:lifecycle}, and \ref{sec:measure} describe actors, lifecycle phases, and measurement in detail. The plane structure encodes two separations that organize the architecture. Within the simulation plane, the simulated platform environment is kept separate from the campaign model studied inside it. Across planes, public platform activity is kept separate from measurement and evidence. Both separations matter because an influence campaign involves far more than content generation. A message must be created, placed into a platform, made visible, encountered by civilians, recorded as an exposure when measurement criteria are met, interpreted by a measurement procedure, possibly translated into a configured state update, and compared against a matched baseline.
\FloatBarrier

The \textbf{control plane} contains the \textit{run controller}, which orchestrates execution across all other components. It initializes the experiment from a configuration and seed, advances discrete simulation steps, schedules actor actions, evaluates the transition conditions summarized in Table~\ref{tab:lifecycle-phase-semantics}, advances the campaign lifecycle when they are met, and records checkpoints and run metadata.

The \textbf{simulation plane} contains the environment, the actors, and the study models, connected through the observe, act, and govern relations shown in the figure. The \textit{environment} is the complete simulated platform in which platform-facing actors operate. It holds the public platform state, including the social graph, content, and discovery surfaces, together with the services that maintain, update, and expose that state. In the reported implementation, OASIS provides accounts, relations, posts, comments, likes, feeds, traces, and platform-compatible timestamps as part of this environment~\cite{yang2024oasis}.

\textit{Actors} are stateful agents that observe the environment through a limited platform view and act on it through validated actions. Section~\ref{sec:env_and_actors} defines three actor categories. Civilians and IO operators are platform-facing actors: civilians form the measured audience, while IO operators act through public platform accounts in active conditions. The manager is a non-public coordination actor that guides IO operators without acting directly on the platform.

The \textit{study models} govern this activity and define what is studied inside the environment. They comprise a campaign model for objectives and lifecycle phases, an action model for permitted behavior, and an exposure model that determines which civilian encounters enter measurement. The study also defines the civilian variables to be measured, which we call constructs. A construct captures a simulator-scale civilian state, such as trust in a source or support for a policy. Each civilian holds an individual value for each construct, and these values can later be aggregated across the measured population. A construct-update rule is configured before a run and applied deterministically within it. The rule determines whether, and by how much, a measured exposure changes an individual civilian's construct value. Section~\ref{sec:measure} defines it formally.

The \textbf{evaluation plane} receives what the simulation plane produces. Its \textit{measurement layer} processes civilian encounters with visible content that satisfy the configured measurement criteria. It records each qualifying encounter as an exposure event, obtains a structured judge reading, and applies the construct-update rule, which may produce either zero or nonzero change in the civilian's measured state. The \textit{evidence layer} preserves links among action records, exposure records, state trajectories, metrics, and matched comparison artifacts. Together, these provenance links allow an aggregate result to be traced back to the exposures, measurements, and platform actions from which it was derived.

The planes meet in the simulation step. At each step, selected actors receive a limited platform view, the action model defines which actions are available, and each selected actor proposes one action. The run controller validates the proposal against actor permissions, lifecycle rules, platform visibility, and study constraints. If approved, the environment applies the action and updates the public platform state. Platform-state, measurement, and evidence flows remain distinct throughout, where approved actions update the public platform state, qualifying civilian encounters enter measurement, and linked provenance records accumulate in the evidence layer. This separation supports controlled comparison and sensitivity analysis, since researchers can vary platform assumptions while holding the study models and measurement rules fixed, or vary those models and rules while holding the environment fixed.
\subsection{Temporal Structures}

IO Factory distinguishes three temporal structures that make simulation time interpretable. 
The \textit{simulation-run lifecycle} describes the execution of one experiment, from configuration and initialization through simulation steps and checkpoints to evaluation and comparison. 
The \textit{campaign lifecycle}, detailed in Section~\ref{sec:lifecycle}, describes the staged influence process being simulated. 
The \textit{actor execution cycle}, described in Sections~\ref{sec:env_and_actors} and~\ref{sec:lifecycle}, names the repeated loop introduced above, in which a platform-facing actor observes a limited platform view, proposes an allowed action, passes validation, commits the action or remains inactive, and leaves a record.

These structures are nested rather than parallel. One simulation run contains many steps, each campaign phase occupies one or more of those steps, and platform-facing actors may complete several execution cycles within a single phase. This nesting keeps simulation steps, campaign phases, and actor actions distinct when interpreting platform state, measured exposure, construct movement, and comparison evidence.

Figure~\ref{fig:simulation-time} shows the timing relation used in the reported implementation. The run controller advances discrete simulation steps, and the IO Factory simulation step is the authoritative temporal index. It labels simulation records and anchors campaign progression, with phases advancing when the transition conditions summarized in Table~\ref{tab:lifecycle-phase-semantics} are met. OASIS provides backend-compatible platform timestamps, while it does not drive lifecycle progression.
\subsection{From Action to Evidence}

Following one message and its associated records through the system shows how the components interact.
An IO operator may propose an action to publish a message. If the proposal passes validation, the environment applies the action and stores the message as a post in the public platform state.

A feed or discovery rule may make the post visible in a civilian's platform view. Visibility alone does not mean that the civilian encountered the post.
If the civilian encounters the post through a configured feed or discovery path and the encounter meets the measurement criteria, the measurement layer creates an exposure record.

Each exposure record then receives a judge reading, a structured simulator measurement that should not be interpreted as ground truth~\cite{zheng2023judging}. The reading supplies inputs to the deterministic construct-update rule described in Section~\ref{sec:measure}, which changes the civilian's construct value only when the configured update conditions are met and yields zero movement otherwise. The evidence layer preserves linked records across the path from public action through exposure and measurement to the update outcome.

After measurement, the campaign can continue, revise guidance, and adapt later activity to the state of the run, constrained by actor roles, lifecycle phase, platform visibility, and action-validation rules~\cite{olejnik2025propagandaFactories}. After the matched runs are complete, civilian construct trajectories from the active run are compared with those from the matched baseline. This comparison yields the simulator-scale directional lift defined in Section~\ref{sec:measure}; it does not estimate real-world persuasion.

\section{Simulated Environment and Actors}\label{sec:env_and_actors}
The reported implementation uses a simulated social-platform environment, summarized in Figure~\ref{fig:system-overview}. The environment stores platform accounts, the follow graph, posts, comments, likes, feeds, discovery surfaces, and platform traces. IO Factory uses this environment as the public world in which actors operate. Campaign rules, actor permissions, exposure rules, and measurement rules are IO Factory instrumentation layered on top of that world; they are constrained by platform state but remain separate from the platform substrate.

The system distinguishes three actor categories.
\begin{enumerate}
    \item \textbf{Civilians} represent ordinary simulated users and form the measured audience. Each civilian receives a platform account, a seeded persona, and a position in the follow graph. At each step, civilians see only the platform content made visible by feed, search, or discovery rules. When selected to act, their next action is generated from that visible context and persona. Study outcomes are measured as changes in civilian simulator-scale construct values, as described in Section~\ref{sec:measure}.
    \item \textbf{IO operators} are controlled intervention actors used in active conditions. They appear on the platform as public accounts, while their internal role gives them access to private campaign guidance and phase-specific intervention actions. They can post, comment, react, follow, search, or otherwise act when the lifecycle phase permits it. Their actions can change what civilians later see.
    \item \textbf{The manager} is a non-public coordination component. It receives structured summaries of campaign state, lifecycle phase, recent platform activity, construct movement, and available opportunities. It issues bounded directives through private coordination channels. These directives affect public behavior only through later IO-operator actions.
\end{enumerate}

Actor state combines stable configuration and mutable simulation state. Stable fields include identity, role, persona, platform account, and action capabilities. Mutable fields include recent activity, visible context, exposure history, source-trust state, susceptibility and resilience parameters, and construct values. This state drives action selection and leaves records for later inspection.

Platform visibility determines whether content can enter measurement. In the directed follow graph, if actor \(B\) follows actor \(A\), then \(A\)'s content can enter \(B\)'s feed. Actors may also encounter content through search, trends, recommendations, or other configured discovery paths. For the reported measurements, the relevant path is whether content reaches civilians, because civilians are the outcome population. A post therefore has no direct effect merely because it exists. It must become visible to a civilian and then be recorded as an exposure before it can enter the measurement pipeline.

The follower graph, feed rule, discovery rule, and interaction model are configurable parts of the experiment. This allows the same campaign model to be tested under different assumptions about platform visibility and audience contact.
\section{Campaign Lifecycle and Intervention Model}\label{sec:lifecycle}

\begin{figure}[!htbp]
  \centering
  \includegraphics[width=\textwidth]{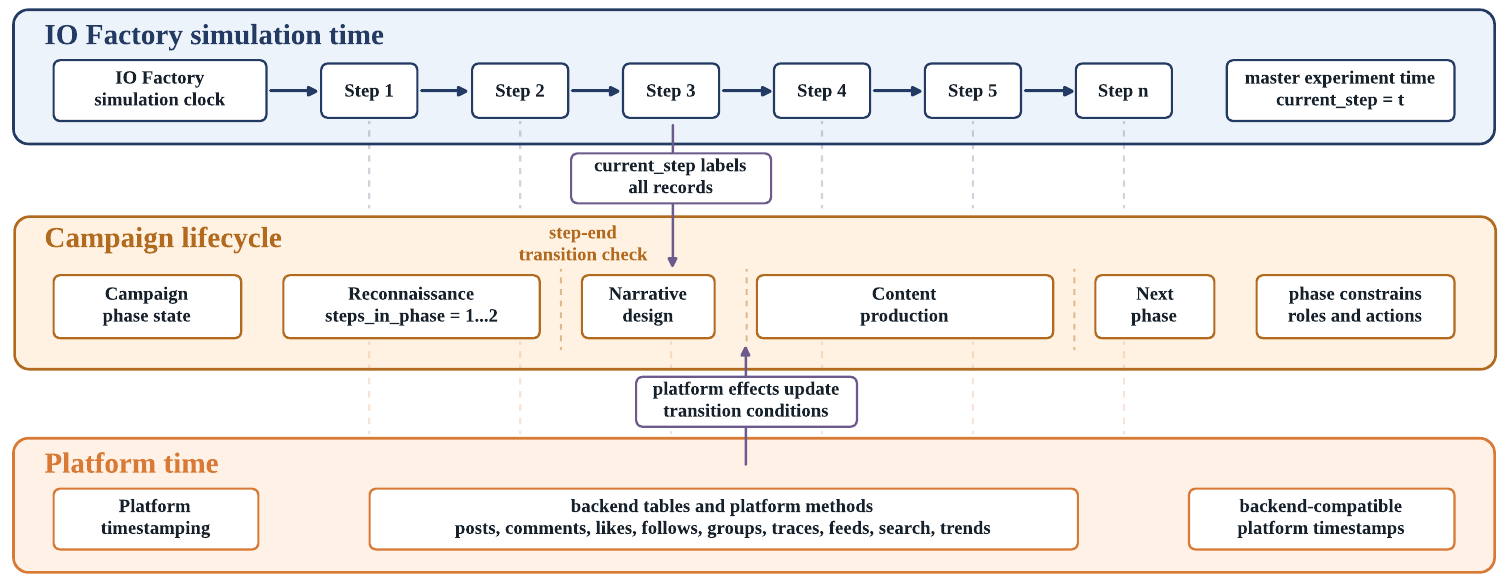}
  \caption{Simulation time coordination. IO Factory advances the authoritative simulation step; campaign phases span multiple steps and change at transition checks. OASIS supplies platform-compatible timestamps but does not drive the campaign lifecycle.}
  \label{fig:simulation-time}
\end{figure}

The campaign lifecycle gives simulated campaign time an operational structure. IO Factory can implement different operational models; the reported study uses a ten-phase reference framework grounded in staged models of information operations and adversarial processes~\cite{cset2021richdata,carnegie2023phase,disarmframework,microsoft2022cyberinfluence,hutchins2011killchain}. The phases are reconnaissance, narrative design, infrastructure, content production, laundering, integration, amplification, absorption, adaptation, and evaluation. As shown in Figure~\ref{fig:simulation-time}, phases span multiple simulation steps and change only when transition conditions are checked.
\FloatBarrier

At each simulation step, the current campaign phase is stored as part of the run state. It gates which actor categories can be selected, which actions they may propose, and which phase-specific records are available for later measurement and comparison. Baseline conditions follow the same phase sequence and measurement rules but contain no IO operators or manager guidance. Table~\ref{tab:lifecycle-phase-semantics} summarizes the operational role of each phase.
\begin{table}[!htbp]
  \centering
  \footnotesize
  \caption{Lifecycle phases and transition gates in the reference implementation. Civilians may continue ordinary platform activity unless noted. Nonterminal phases require minimum steps; maximum-step caps can force advancement.}
  \label{tab:lifecycle-phase-semantics}
  \begin{tabularx}{\textwidth}{@{}>{\raggedright\arraybackslash}p{0.14\textwidth}
  >{\raggedright\arraybackslash}p{0.25\textwidth}
  >{\raggedright\arraybackslash}p{0.24\textwidth}
  >{\raggedright\arraybackslash}X@{}}
    \toprule
    \textbf{Phase} & \textbf{IO / manager activity} & \textbf{Operational role} & \textbf{Transition gate after minimum dwell} \\
    \midrule
    Reconnaissance & IO operators analyze the simulated population; no public IO posting. & Builds population, network, construct, and susceptibility summaries. & Reconnaissance marked complete. \\
    Narrative design & Manager and IO operators generate narrative candidates; no public IO posting. & Produces internal campaign objects for later action. & At least one narrative is ready, or narrative design is bypassed in a silent-control run. \\
    Infrastructure & IO operators establish platform presence through allowed non-campaign actions. & Establishes accounts, graph position, feed access, and coordination conditions. & Infrastructure marked deployed. \\
    Content production & IO operators may publish initial campaign-relevant posts or comments. & First phase in which campaign material enters the public platform. & Content-production threshold is met, or content production is bypassed in a silent-control run. \\
    Laundering & IO operators may relay or cross-reference existing campaign material. & Records relay and source-path movement when configured. & Eligible relay-action threshold is met; gate is skipped when no amplifier actors are configured. \\
    Integration & IO operators may engage civilian-visible conversations and accounts. & Places campaign material into broader platform contexts. & Reach threshold is met; baseline runs use a fixed dwell because no campaign operators are present. \\
    Amplification & IO operators continue full public platform activity. & Supports repeated circulation, replies, shares, and broader reach. & Minimum amplification dwell is complete. \\
    Absorption & IO operators are not the active public driver; civilians may encounter circulating material. & Records exposure, judge readings, possible state updates, and delayed movement. & Minimum absorption dwell is complete. \\
    Adaptation & Manager assesses progress and may issue revised guidance. & Uses recorded state to continue, revise, loop back, or proceed to evaluation. & Goal met, loop-back target selected, or maximum adaptation loops reached. \\
    Evaluation & IO operators do not publish campaign content. & Produces terminal summaries, trajectories, inclusion status, and audit records. & Terminal phase. \\
    \bottomrule
  \end{tabularx}
\end{table}

\FloatBarrier
\subsection{Narrative Formation}
Narrative formation translates configured construct definitions into intermediate campaign objects. In the reported implementation, the narrative-design phase uses the configured language model to generate structured narrative candidates from each target construct and target direction. Each candidate records a core message, emotional hooks, keywords, content templates, and plausibility or effectiveness proxies. These objects guide later IO-operator actions, but they are not public platform content and are not outcome measures.

A narrative candidate can shape later action selection or text generation, but it cannot directly change civilian construct state. Civilian movement can occur only after a later IO-operator action is validated, committed to the platform, made visible to a civilian, recorded as an exposure, interpreted by the judge, and passed through the construct-update rule described in Section~\ref{sec:measure}. IO Factory therefore records narrative formation as one step in the evidence path from construct definition to platform-visible content and possible measured state change.
\subsection{Actor Execution and Manager Guidance}
After initialization, platform-facing actors follow an observe, select, validate, commit, and record cycle that is phase-gated. At each step, the run controller selects eligible actors according to the run configuration, activity schedule, and current campaign phase. Each selected actor receives a limited platform view, such as recent posts from followed accounts, search results, trend results, or other configured discovery outputs. IO Factory then builds the action-selection context from this visible platform state, the actor persona, current constraints, and any private guidance available to that actor.

An LLM call proposes one action from the allowed action set. If the action requires text, such as a post, comment, or quote, a separate generation call produces the content. Validation separates proposed behavior from platform effect: IO Factory executes the action only if it satisfies actor permissions, lifecycle rules, platform visibility, and study constraints. Once validated, the action is committed to the platform and added to the simulation record.

Manager directives enter the actor execution cycle as private guidance for eligible IO operators. The manager receives campaign status, including phase history, active narratives, opportunity signals, and recent metrics. Its output can shape later IO-operator action selection, but any public effect must still pass through IO-operator action selection, content generation where needed, validation, platform commitment, visibility, and measurement. The manager therefore coordinates the intervention without directly creating public platform content.

\section{Measurement Pipeline and Comparison Protocol}\label{sec:measure}
The measurement pipeline connects platform activity to possible movement in simulated civilian state. It follows the path introduced in Figure~\ref{fig:system-overview}: content is created, stored in the platform environment, made visible through discovery rules, encountered by a civilian, evaluated against a construct definition, and then passed to a deterministic update rule. A construct definition is the study-specific description of what is being measured, such as trust in a source or support for a policy. It specifies the construct scale, target direction, relevance criteria, and the type of content that can count as moving the construct.

For example, the public-institutions construct measures trust in public-sector institutions on the simulator scale. Its high pole describes such institutions as competent, legitimate, fair, and acting in the public interest; its low pole describes them as corrupt, captured, incompetent, or abusive. Content is relevant only when it is materially about public-sector institutions, and in the decrease-target study relevant content pointing toward the low pole can reduce the simulated construct value through the update rule.

IO Factory keeps message production, visibility, exposure, interpretation, and state update separate. This distinction is central to the framework. A post may be generated but never become visible to a civilian. It may become visible but not be selected into measurement. It may be measured but judged irrelevant to the construct. It may be relevant but too weak, insufficiently confident, or in the wrong direction to change the simulated state. Recording these stages separately prevents campaign volume from being confused with exposure or effect.

An exposure event is the measurement record created when a civilian encounters a platform object through a configured visibility or discovery path. When that event enters measurement, an LLM judge evaluates the exposed content against the construct definition. The judge returns structured readings for relevance, stance, confidence, and persuasiveness. These readings are simulator measurements, not ground truth. They parameterize the update rule; the update itself is deterministic given the exposure record, judge output, civilian state, and experiment configuration.

For civilian \(i\), construct \(c\), and exposure event \(e\), the simulator-scale movement is
\begin{equation*}
\Delta_{i,c,e}
=
p_e \, \tau_{i,s(e)} \, u_i \, r_{i,e} \, \lambda_c \, d_{e,c} \, \gamma_{e,c}.
\label{eq:construct-update}
\end{equation*}

The civilian state is then updated by

\[
\begin{aligned}
x_{i,c}^{\,new}
&=
\operatorname{clip}
\left(x_{i,c}^{\,old} + \Delta_{i,c,e}\right),\\
\operatorname{clip}(z)&=\min\{10,\max\{0,z\}\}.
\end{aligned}
\]

Here \(p_e\) is the judged persuasiveness of the exposure, \(\tau_{i,s(e)}\) is civilian \(i\)'s trust in the source of exposure \(s(e)\), \(u_i\) is a civilian-level update weight, \(r_{i,e}\) is the repetition weight for repeated exposure to the same content, \(\lambda_c\) is the construct-specific update rate, \(d_{e,c}\) is the direction of the exposure with respect to construct \(c\), and \(\gamma_{e,c}\) is the judge confidence for that construct reading. The clipping operation keeps simulated construct values on the configured scale.

The update rule reflects a simple premise from persuasion and misinformation research: message volume alone does not determine effect. Source trust, repetition, prior belief, familiarity, and processing route can all shape interpretation~\cite{petty1986elaboration,chaiken1980heuristic,ecker2022psychological,dechene2010truth}. IO Factory makes these assumptions explicit by placing them in the configured update rule. The framework therefore does not hide persuasion assumptions inside the language model; it records which factors were allowed to affect simulated state.

Repeated exposure is handled by a configurable decay term. For repeated exposure by civilian \(i\) to the same content item, the repetition weight is
\[
r_{i,e} = \max(\rho_{\min}, \rho^{n_{i,e}}),
\]
where \(n_{i,e}\) is the number of prior processed exposures by civilian \(i\) to that content item, \(\rho\) is the decay factor, and \(\rho_{\min}\) is the minimum repetition weight. This allows repetition to matter without assuming that every repeated encounter has the same force as the first one.

The comparison protocol separates ordinary simulator dynamics from the effect of the configured intervention. For each active run, IO Factory uses a matched baseline run with the same civilian population, platform setup, construct definitions, exposure processing, update rule, and outcome calculation. The matched baseline run does not include IO operators or manager guidance. The active run adds those intervention components while leaving the rest of the setup fixed. In both runs, outcomes are computed over civilians only. The comparison therefore asks how much farther civilian construct values move when intervention actors are present than when the same simulated population evolves without them.

For seed \(s\) and construct \(c\), directional lift is computed as
\[
L_{s,c}
=
q_c
\left[
\left(\bar{x}^{\mathrm{active}}_{s,c,T}-\bar{x}^{\mathrm{active}}_{s,c,0}\right)
-
\left(\bar{x}^{\mathrm{baseline}}_{s,c,T}-\bar{x}^{\mathrm{baseline}}_{s,c,0}\right)
\right].
\]
The sign \(q_c\) aligns the result with the target direction of the construct. For increase-target constructs, \(q_c=1\). For decrease-target constructs, \(q_c=-1\). Positive directional lift therefore means movement in the intended campaign direction after subtracting matched baseline movement. Section~\ref{sec:simulation-results} reports these comparisons for the empirical runs.
\section{Simulation Runs and Results}\label{sec:simulation-results}
We evaluate IO Factory as an end-to-end framework for simulating AI-enabled influence campaign lifecycles in a controlled platform environment. The main question is whether the framework can execute active campaign behavior and produce interpretable evidence about exposure, construct movement, and platform circulation.

\paragraph{Study design}
The primary comparisons use matched baseline and active runs. Baseline runs retain the civilian population, platform mechanics, exposure processing, judge-based measurement, and construct-update rules. Active runs use the same setup and add IO operators, manager guidance, and phase-specific intervention actions. Each main design uses 13 matched simulation replicates with 10,000 civilians per condition; active runs add 1,000 IO operators. Inference is conducted over matched simulation replicates, not over civilians, posts, exposures, graph edges, or steps. The reported runs use Gemma 4 31B on H200 nodes; the model is a configuration choice and can be replaced without changing the framework architecture.

We report two main empirical designs. The first is a two-construct increase-target design that targets \textit{trust in Russia} and \textit{support for eating insects} in the same active lifecycle. The second is a single-construct decrease-target design that targets lower \textit{trust in public institutions}. The designs are interpreted separately because they differ in construct load, scenario content, and target direction. We also conduct a large-scale validation run with 100,000 civilians and 10,000 IO operators to demonstrate execution at that scale; the matched 10,000-civilian designs provide the evidence about movement relative to baseline.
\begin{table}[!htbp]
\centering
\scriptsize
\setlength{\tabcolsep}{3pt}
\renewcommand{\arraystretch}{1.06}
\caption{Civilian-state diagnostics after the matched baseline and active runs. Values are mean $\pm$ SD across the 13 matched replicates on the simulator scale. Primary construct outcomes and directional lift are reported in Figure~\ref{fig:directional-lift-trajectories}.}
\label{tab:construct-roles-and-movement}

\begin{tabularx}{\textwidth}{@{}
>{\raggedright\arraybackslash}p{2.75cm}
>{\raggedright\arraybackslash}X
>{\centering\arraybackslash}p{2.15cm}
>{\centering\arraybackslash}p{2.15cm}
@{}}
\toprule
\textbf{Measure} &
\textbf{Meaning} &
\textbf{Baseline} &
\textbf{Active} \\
\midrule

\multicolumn{4}{@{}l}{\textbf{Two-construct increase-target design}}\\
\midrule

Polarization &
Diagnostic (dispersion in civilian construct values). &
1.906 $\pm$ 0.013 &
2.001 $\pm$ 0.026 \\

Follow-edge similarity &
Diagnostic (mean construct-state similarity across civilian follow edges). &
0.843 $\pm$ 0.001 &
0.839 $\pm$ 0.001 \\

\midrule
\multicolumn{4}{@{}l}{\textbf{Single-construct decrease-target design}}\\
\midrule

Polarization &
Diagnostic (dispersion in civilian construct values). &
3.865 $\pm$ 0.036 &
4.714 $\pm$ 0.349 \\

Follow-edge similarity &
Diagnostic (mean construct-state similarity across civilian follow edges). &
0.776 $\pm$ 0.002 &
0.750 $\pm$ 0.010 \\

\bottomrule
\end{tabularx}
\end{table}

\FloatBarrier

\begin{figure}[!htbp]
\centering
\includegraphics[width=\linewidth]{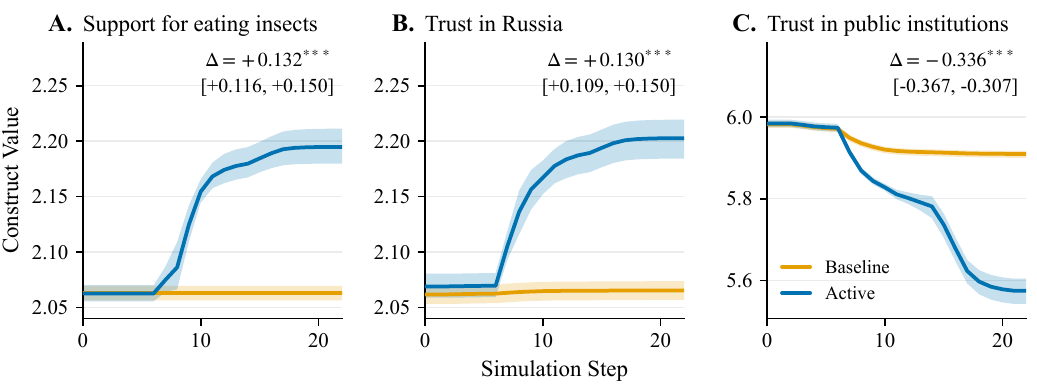}
\caption{Mean construct value trajectories under matched baseline and active conditions. Lines average the 13 matched replicates and ribbons show bootstrap 95\% intervals. Panels A and B share one active run and one scale, and panel C shows the separate decrease-target run on its own scale. $\Delta$ is the final-step active-minus-baseline change from run start, shown with its paired-bootstrap 95\% interval and exact sign-flip significance after Holm adjustment ($^{***}\,p<0.001$). In panel C, $\Delta=-0.336$ corresponds to a sign-adjusted directional lift of 0.336.}
\label{fig:directional-lift-trajectories}
\end{figure}

\paragraph{Primary construct movement}
Figure~\ref{fig:directional-lift-trajectories} reports primary construct movement, and Table~\ref{tab:construct-roles-and-movement} reports accompanying civilian-state diagnostics. Primary movement is summarized using the directional-lift measure defined in Section~\ref{sec:measure}. Confidence intervals are paired-bootstrap intervals over the 13 matched baseline-active replicates; reported \(p\)-values use exact paired sign-flip tests with Holm adjustment across the three primary construct endpoints.

All three primary endpoints move in the target direction. In the two-construct increase-target design, directional lift is 0.132 for support for eating insects and 0.130 for trust in Russia. In the single-construct decrease-target design, trust in public institutions declines more than baseline, yielding sign-adjusted lift of 0.336. All three directional-lift estimates are significant after Holm correction, with \(p<0.001\).
\FloatBarrier

\paragraph{Exposure, phase, and discourse diagnostics}
\begin{wrapfigure}[23]{r}{0.50\textwidth}
\centering
\includegraphics[width=0.98\linewidth]{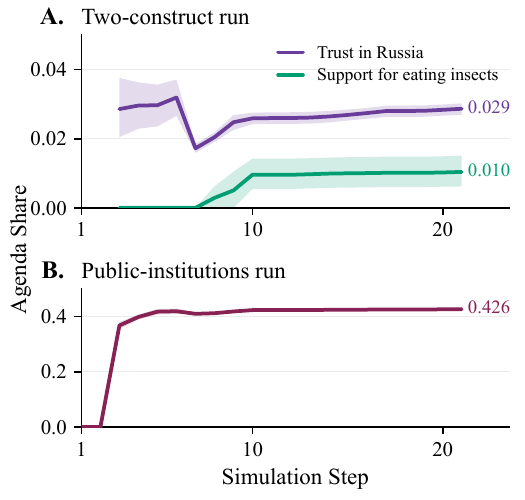}
\caption{Agenda share in measured civilian-authored posts under active conditions. Panels show the two-construct run (A) and the public-institutions run (B). Lines average 13 matched replicates with bootstrap 95\% ribbons and final topic-share labels.}
\label{fig:agenda-diagnostics}
\end{wrapfigure}

Figure~\ref{fig:agenda-diagnostics} reports target-topic presence in measured civilian-authored posts. Agenda share is the share of measured civilian-authored posts at a step judged as containing the target topic~\cite{mccombs1972agenda}. Baseline topic-presence estimates are not plotted because the reported baseline configurations did not run target-topic agenda accounting. The agenda measure is therefore a simulator diagnostic of discourse uptake, not independent validation of construct movement.

Exposure records describe how platform activity entered the construct-update path. A processed exposure event is an exposure record evaluated for possible construct movement. The update rate is the share of processed exposures that changed construct value. Source trust, persuasiveness, stance confidence, repetition weight, and mean update size are simulator variables used by the deterministic update rule; they are not external causal estimates.

Similar final lift can arise through different exposure and update profiles. The two increase-target constructs had comparable processed-exposure counts, about \(495k \pm 71k\), but different update rates: \(29.3\%\) for support for eating insects and \(33.5\%\) for trust in Russia. The single-construct decrease-target design had more processed exposures, \(597k \pm 66k\), a higher update rate of \(90.4\%\), and a larger mean update size. Phase summaries localize when movement occurred within the lifecycle: in the two-construct design, movement is concentrated in content production; in the single-construct design, the largest movement appears during amplification, followed by content production.

Final active-condition agenda share was \(0.039 \pm 0.009\) in the two-construct design, with \(0.029 \pm 0.003\) for the trust-in-Russia topic and \(0.010 \pm 0.008\) for the insect-eating topic. The public-institutions design reached \(0.426 \pm 0.005\). The two-construct result shows why construct movement and civilian-authored topic presence are kept separate: trust in Russia and support for eating insects had similar directional lift, but civilian-authored discourse contained more Russia-topic material than insect-eating-topic material.

\paragraph{Platform reach and circulation}
Platform reach and circulation measures describe contact and spread, not construct movement. Table~\ref{tab:delivery-circulation} reports IO-operator reach, civilian-mediated reach, engagement, civilian relay activity, and coordinated posting in active conditions.
\begin{table}[!htbp]
\centering
\footnotesize
\caption{Platform reach and circulation at the final step. Values are mean $\pm$ SD across active runs. Measures are platform diagnostics, not primary construct endpoints.}
\label{tab:delivery-circulation}
\setlength{\tabcolsep}{3pt}
\begin{tabularx}{\textwidth}{@{}
>{\raggedright\arraybackslash}p{4.9cm}
>{\centering\arraybackslash}p{2.15cm}
>{\centering\arraybackslash}p{2.25cm}
>{\centering\arraybackslash}p{2.00cm}
>{\centering\arraybackslash}p{2.25cm}
>{\centering\arraybackslash}p{2.35cm}
@{}}
\toprule
\textbf{Empirical design} & \textbf{IO-operator reach fraction} & \textbf{Civilian-mediated reach} & \textbf{Engagement rate} & \textbf{Civilian relay activity} & \textbf{Coordinated posting score} \\
\midrule
Two-construct increase-target & 0.494 $\pm$ 0.003 & 9267 $\pm$ 70 & 2.009 $\pm$ 0.421 & 0.154 $\pm$ 0.375 & 1.000 $\pm$ 0.000 \\
Single-construct decrease-target & 0.494 $\pm$ 0.003 & 9391 $\pm$ 44 & 1.759 $\pm$ 0.362 & 0.002 $\pm$ 0.001 & 1.000 $\pm$ 0.000 \\
\bottomrule
\end{tabularx}
\end{table}

\FloatBarrier

The active reach fraction is about 0.494 in both main designs, meaning roughly half of the modeled civilians had an interaction record involving an IO-operator source. Civilian relay activity is measurable but low under the configured metric, especially in the single-construct design. These measures help separate platform contact and relay from the construct movement reported in Figure~\ref{fig:directional-lift-trajectories}.

\paragraph{Summary}
Together, the results show that IO Factory can execute campaign-lifecycle simulations, compare active and baseline runs, and trace construct movement through exposure records, judge readings, population-state diagnostics, and circulation records. These are simulator results under declared assumptions. They do not estimate real-world persuasion or operational effectiveness.

\section{Discussion and Roadmap}\label{sec:discussion}
IO Factory introduces a framework for representing AI-enabled influence campaigns as inspectable lifecycle simulations. The reported runs show that the system can generate campaign activity, route it through a simulated platform, record civilian exposure, apply declared measurement rules, and compare active runs against matched baselines. The contribution is not a claim about real-world persuasion. It is a way to make campaign assumptions, coordination paths, exposure records, and measurement rules explicit.

This has implications for informational resilience and threat-intelligence analysis. Campaigns built from AI agents may not be identifiable from single posts or single accounts. They may appear through persistent personas, repeated low-intensity interactions, semantic coordination without text reuse, source laundering, or adaptation after feedback~\cite{schroeder2026swarms,olejnik2025propagandaFactories,kunst2026cyborg,orlando2026emergent,ferrara2016rise,pacheco2021uncovering,tardelli2024temporal,mannocci2024detection,stratcom_laundering}. IO Factory addresses this gap by turning campaign-level patterns into controllable simulation objects before they are observed at scale on real platforms. The simulator does not establish field effectiveness, but it can make assumptions explicit enough for defensive exercises, red-teaming, and threat-model development.

The results should be interpreted within three boundaries. First, the reported runs use modeled civilians, configurable platform rules, a simulated exposure graph, and construct movement on a simulator scale. These results therefore should not be read as estimates of real-world persuasion. External claims require calibration, sensitivity analysis, and comparison with real observations where such comparison is ethical and possible~\cite{park2023generative,gao2024large,argyle2023out,aher2023using}.

Second, LLM-based measurement makes large-scale construct measurement feasible, but LLM judges are not ground truth~\cite{liu2023geval,zheng2023judging,bowman2021benchmarking}. Their outputs depend on construct definitions, prompts, model versions, serving settings, and context. Holding judge settings fixed across matched conditions supports internal comparison, but it does not externally validate the judge. Future work should calibrate judge outputs against human annotations and report sensitivity to judge prompts and model choice.

Third, the reported results are model-dependent. LLMs are used for narrative design, action selection, content generation, manager assessment, and judge measurement. IO Factory records model identity, settings, prompts, and model-call provenance, but the reported results do not establish that actor behavior or narrative formation is invariant across model families. Preliminary runs with Gemma, Qwen, and GLM models show that the framework can execute with different models, while narrative specificity and downstream content formation may vary. Cross-model sensitivity analysis should therefore become a standard part of future IO Factory studies.

Future work should develop IO Factory as shared research infrastructure. Important next steps include benchmark scenarios with common construct definitions, seeds, reporting rules, and artifact schemas~\cite{wilkinson2016fair,gebru2021datasheets,mitchell2019model,raji2020closing}. The framework should also support systematic variation in graph structure, recommender behavior, source trust, repetition, judge prompts, and actor policies. Defensive scenarios should become first-class interventions, including counter-campaigns, platform moderation, inoculation strategies, and cyber-range-style evaluation designs~\cite{perez2022red,yamin2020cyber}.

A further extension is to distinguish human-facing exposure from machine-facing exposure. A human civilian encounters content through feeds, search, recommendations, or direct interaction. A machine agent may encounter content through retrieval, browsing, summarization, recommender input, or later training data. This distinction matters because retrieval corpora and web-scale datasets can themselves become targets for manipulation or poisoning~\cite{zhong2023poisoning,carlini2024poisoning}.

The longer-term goal is a shared representational vocabulary for AI-enabled influence research. The community needs a way to state what was represented, what changed, who was measured, which records count as evidence, and where interpretation stops. IO Factory is a reference implementation of that vocabulary.
\section{Conclusion}\label{sec:conclusion}
Influence operations are processes, not isolated posts or bursts of content. This is why AI-enabled influence campaigns are difficult to study from content alone. IO Factory represents such campaigns as traceable, structured processes inside a controlled social-platform simulation. It separates the platform environment from the campaign model, records how actors act within that environment, follows content through visibility and exposure, and compares measured civilian-state movement against matched baselines.

The reported runs show that IO Factory can execute full campaign lifecycles at simulation scale and preserve an inspectable evidence path from public action and non-public coordination to exposure records, judge readings, deterministic state updates, and outcome summaries. The results are simulator-scale measurements under declared assumptions. They do not estimate real-world persuasion.

The contribution is a framework for studying AI-enabled influence as a systemic framework. IO Factory provides a structured basis for red-team scenario design, benchmark construction, sensitivity analysis, and future empirical work on campaign-level risks from AI-agent coordination.

\bibliographystyle{IEEEtran}
\bibliography{references}

\end{document}